\documentclass[10pt,twocolumn,letterpaper]{article}

\usepackage{wacv}
\usepackage{mystyle}

\def\wacvPaperID{****} 

\wacvalgorithmstrack   

\wacvfinalcopy 

\ifwacvfinal
\usepackage[breaklinks=true,bookmarks=false]{hyperref}
\else
\usepackage[pagebackref=true,breaklinks=true,colorlinks,bookmarks=false]{hyperref}
\fi

\begin{document}

\title{VidConv: A modernized 2D ConvNet for Efficient Video Recognition }

\author{Chuong H. Nguyen, Su Huynh, Vinh Nguyen, Ngoc Nguyen \\
	CyberCore AI, Ho Chi Minh, Viet Nam \\
	{\tt\small {chuong.nguyen,su.huynh,vinh.nguyen,ngoc.nguyen}@cybercore.co.jp}
}

\maketitle
\thispagestyle{empty}

\begin{abstract}
	
	Since being introduced in 2020, Vision Transformers (ViT) has been steadily breaking the record for many vision tasks and are often described as ``all-you-need" to replace ConvNet. Despite that, ViTs are generally computational, memory-consuming, and unfriendly for embedded devices. In addition, recent research shows that standard ConvNet if redesigned and trained appropriately can compete favorably with ViT in terms of accuracy and scalability. In this paper, we adopt the modernized structure of ConvNet to design a new backbone for action recognition. Particularly, our main target is to serve for industrial product deployment, such as FPGA boards in which only standard operations are supported. Therefore, our network simply consists of 2D convolutions, without using any 3D convolution, long-range attention plugin, or Transformer blocks. While being trained with much fewer epochs (5x-10x), our backbone surpasses the methods using (2+1)D and 3D convolution, and achieve comparable results with ViT on two benchmark datasets.                
	
\end{abstract}

\section{Introduction}
Efficient video recognition models are desired for many applications, such as security monitoring, shop-lifting protection, or driver-action recognition, where the trade-off between efficiency and accuracy is critical. Despite recent advances in deep network architecture, selecting a model that can run on embedded devices and achieve high accuracy for video recognition is still challenging. The main building blocks of current methods rely either on 3D convolution (3D ConvNet) or Transformer. They achieve state-of-the-art (SOTA) accuracy by gradually stacking more layers or plugins and being training with lots of extra data. Unfortunately, the price for a little accuracy improvement is often an expensive and much longer training. In addition, while adding novel operators can be easily implemented with popular framework such as TensorFlow and Pytorch on GPU device, optimizing parallel computation unit (PCU) of embedded devices for these operators can be very costly and time-consuming, thus rarely supported in commercial devices. 

\begin{figure}[h]
	\begin{subfigure}[b]{0.49\columnwidth}
		\centering
		\includegraphics[width=1\columnwidth]{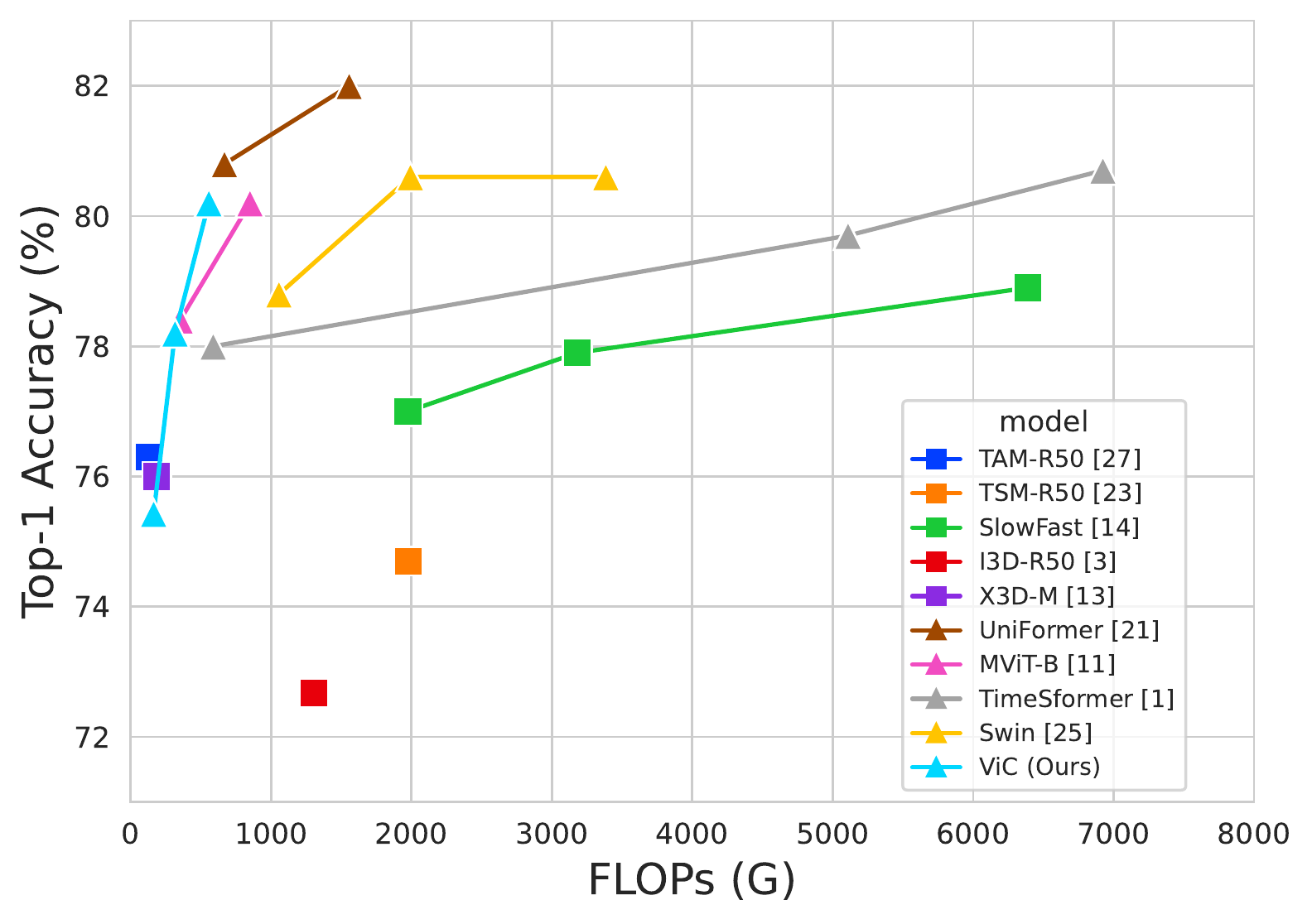}
		\caption{Accuracy \vs FLOPs}
		\label{fig:AccFlop}
	\end{subfigure}%
	\hfill
	\begin{subfigure}[b]{0.47\columnwidth}
		\centering 
		\includegraphics[width=1\columnwidth]{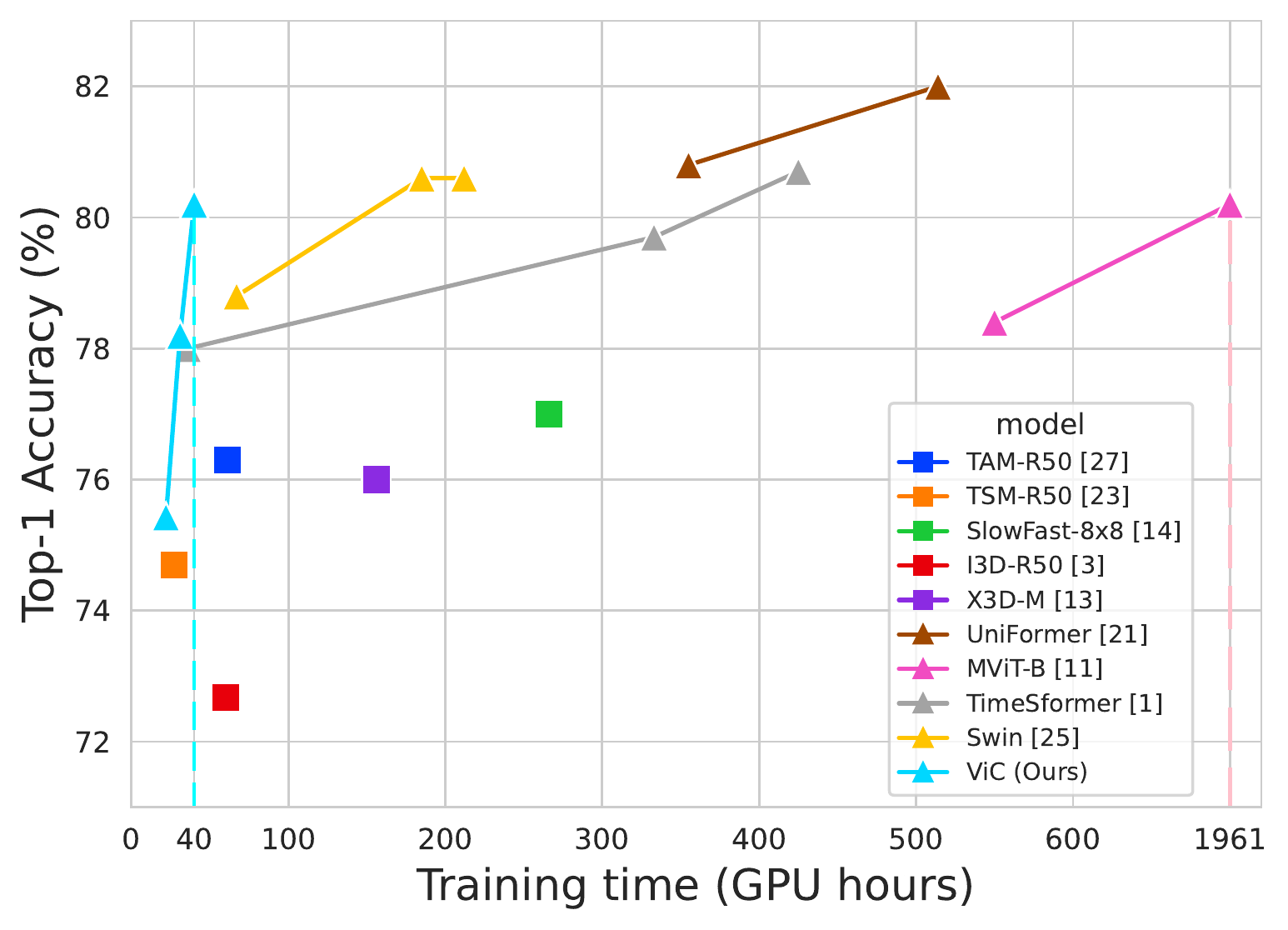}
		\caption{Accuracy \vs Training Time}
		\label{fig:AccTrainingTime}
	\end{subfigure}
	\caption{Comparision of our prosposed VidConv (ViC) with SOTA approaches for action recognition on Kinetics-400. FLOPs is reported for total number of views (more details in \Tab{SOTA-K400}).}
\end{figure}

In this paper, we aim to bridge the gap between achieving good accuracy and being efficient for training and deployment by introducing a new backbone network for video recognition, dubbed as VidConv:  
\begin{itemize}
	\item Our model uses only standard 2D convolution that is fully supported in all embedded devices and deployment library. In particular, we neither inflate a 2D Convolution into 3D version, nor adding self-attention as in Transformer. Instead, we flatten a 3D video tensor $T\times H\times W $ into a 2D tensor $hH \times wW$, where $T=h.w$ is the number of frames in a clip, and $H,W$ are the height and width of a frame, respectively. Then, we use two 2D depth-wise convolutions: one with $7 \times 7$ kernel and another of kernel $(h,w)$ and dilation $(H,W)$ to aggregate spatial and temporal features. This simple design reduces the number of parameters significantly relative to 3D counterpart, and does not consume large memory and computational softmax operator as in attention module.   
	\item As shown in \fig{AccTrainingTime}, our model needs much less training time (5$\times$-10$\times$), thus it is much more cost-effective and suitable for industrial applications.
	\item Efficiency - Accuracy trade off: Since using only 2D convolutions, our model is simple and very efficient. On two benchmark datasets Kinetic-400 and Something-Something, our models surpass all recent methods using (2+1)D or 3D Convolution, and achieve comparable results with Vision Transformer models.  
\end{itemize}  


\section{Related Work}
\subsection{CNN and variant}
The dominant approach for video understanding has been based on 3D Convolution Neural Networks (3D-CNN). Among them, Slow-Fast Network \cite{feichtenhofer2019slowfast} has been considered as the strong baseline for action recognition and other downstream tasks. However, 3D CNN suffer from the large computation cost and difficult optimization. The early works, such as I3D \cite{carreira2017quo}, constructs the 3D CNN by inflating the pretrained 2D convolution models, which help to stabilize the training due to better initialization. To reduce the complexity, later works (\cite{tran2018closer}, \cite{xie2018rethinking}, \cite{tran2019video}) factorize the 3D kernel into (2+1)D convolution in the spatial and temporal dimensions. X3D \cite{feichtenhofer2020x3d} and MoviNet \cite{kondratyuk2021movinets} are developed by Neural Architecture Search (NAS) to optimize the model efficiency. However, convolution networks are recently considered inferior to Transformer-based approaches due to its limited receptive field, thus can't capture the long range temporal information.   

\subsection{Temporal Module plugin}
Other methods design plugin-modules that are inserted into middle layers of the standard 2D networks, such as ResNet, to capture short-term and long-term temporal information. Non-local Network \cite{wang2018non} was proposed to improve the ability to model long range relationship between the frames.  Temporal Shift Module \cite{lin2019tsm} shifts a portion of channels along the temporal axis to fuse the information between the frames. TAM \cite{liu2020tam} propose a two-level module, which decouples the dynamic kernel into a location sensitive map and a location invariant aggregation weight. The former is learned in a local temporal window to capture short-term information, while the later is generated from a global view with a focus on long-term structure. 
Kwon \etal \cite{kwon2020motionsqueeze} propose the Motion Squeeze module that extracts motion cues using frame-wise appearance correlation, and combines the motion feature into the next downstream layer. Similarity, STM \cite{jiang2019stm} and TEA \cite{li2020tea} encode motion information by differentiating adjacent features across frames, which are then combined with the spatial features. However, these methods are convolved along the spatial and temporal axes separately. This limits the interaction between spatial and temporal features compared to 3D spatio-temporal or transformer based methods. Hence, Lee  \etal \cite{lee2021diverse} propose a temporal 3D module that aggregates hierarchy features with different temporal receptive fields. This helps capture the large visual tempo variation of the action. Although these temporal modules can boost the performance of the networks, they often intertwine different branches and connections across the layers and frames, limiting it for deployment on hardware devices.  

\subsection{Vision Transfomer and variant}
Vision Transformer (ViT) (\cite{dosovitskiy2020image}, \cite{touvron2021training}, \cite{liu2021swin}) has been creating a new wave in network architecture design for vision tasks. Motivated from the success of transformer in NLP, ViT divides an image into many patches, which is analogous to visual words and then applies the vanilla attention mechanism \cite{vaswani2017attention} on these patches to extract the features. Extending the ``2D-patchifying" approach into 3D domain for video, several recent works successively propose different variants for spatiotemporal learning, such as VTN \cite{sharir2021image}, STAM \cite{neimark2021video}, TimeSformer \cite{gberta_2021_ICML}, Video Swin-Transformer \cite{liu2021video} and MultiScale ViT \cite{fan2021multiscale}. Self-Attention mechanism is believed to be the key for success of ViT thanks to its ability to capture the long-range relationship. However, ViT models are often memory consuming and computational. In addition, recent research \cite{liu2022convnet}, \cite{yu2021metaformer}, \cite{tolstikhin2021mlp} show that simple MLP and Convolution networks, if trained in a similar recipe, can compete favorably with ViT, questioning the true effectiveness of ViT.  

Therefore, in this work, our objective is not to introduce a new or complex temporal module. Instead, we select a minimal design of the 2D ConvNet, that is typically considered inferior to 3D ConvNet or ViT for action recognition, and prove that its performance is still very competitive while preserves the efficiency. We hope that our model can serve as a simple and strong baseline, and offer a practical solution for product deployment.  
\section{VidConv Model}
\subsection{Overall Architecture}
Our VidConv is built upon the recent modernized ConvNet, particularly the ConvNeXt \cite{liu2022convnet} thanks to its high performance and simplicity, but other variants such as PoolFormer \cite{yu2021metaformer} are still applicable.  

\begin{figure*}[t]
	\centering
	\includegraphics[width=0.95\textwidth]{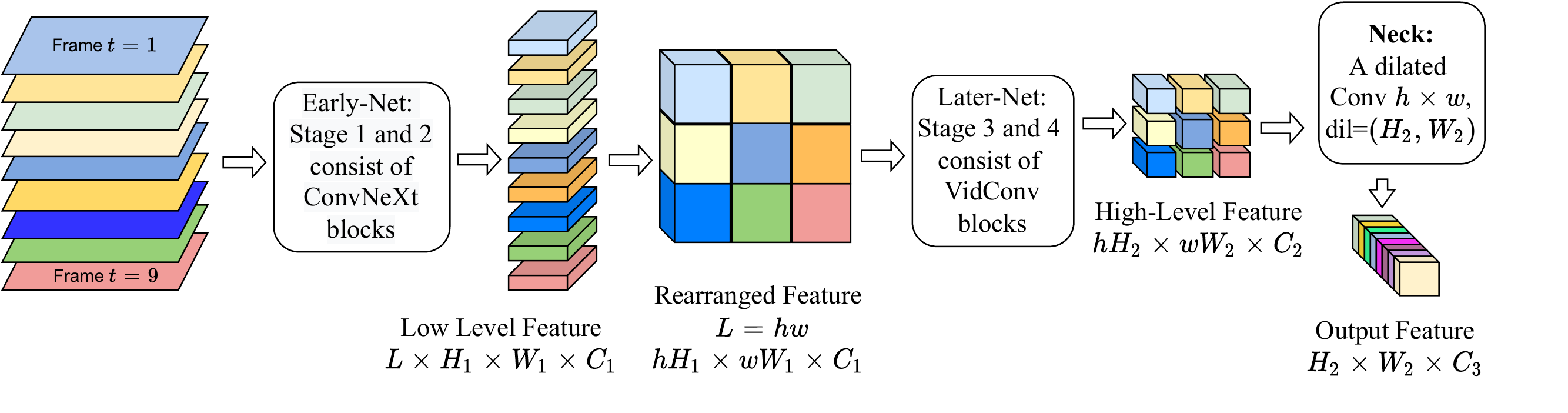}
	\caption{Illustration of VidConv Network with $L=9$ frames. We stack low level features from 3D dimension $(L \times H_1\times W_1)$ into 2D feature of spatial dimension $(h H_1 \times w W_1)$, where $h=w=3$, $(H_1,W_1)=(H/8,W/8)$ and $(H_2,W_2)=(H/32,W/32)$.}
	\label{fig:VidConvPipeline}
\end{figure*}
\begin{figure*}[t]
	\centering
	\begin{subfigure}[b]{0.45\textwidth}
		\centering
		\includegraphics[width=1\columnwidth]{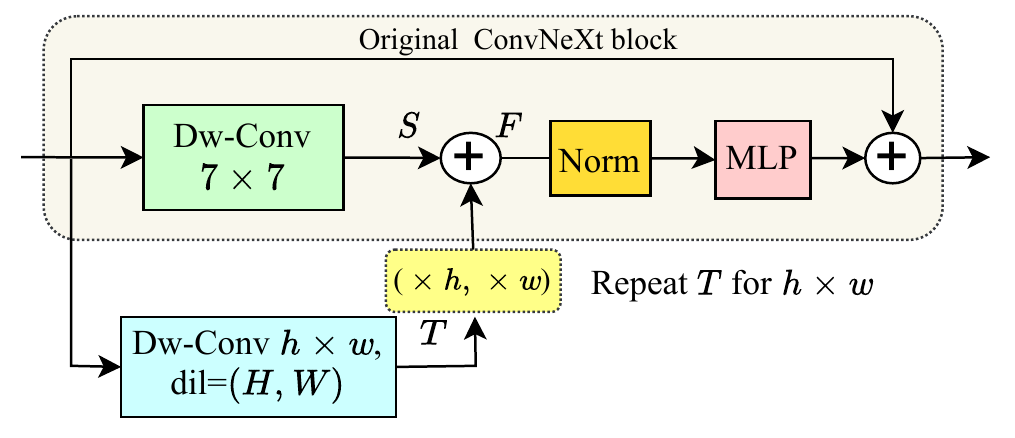}
		\caption{VidConv Block Diagram}
		\label{fig:VidConvBlock}
	\end{subfigure}
	\hspace{1.5cm}
	\begin{subfigure}[b]{0.28\textwidth}
		\centering
		\includegraphics[width=1\columnwidth]{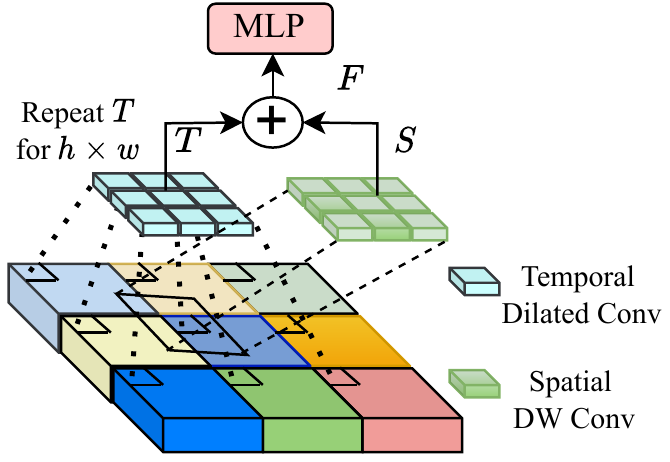}
		\caption{Visualize VidConv Operation}
		\label{fig:VidConv3DBlock}
	\end{subfigure}
	\caption{VidConv block diagram and its operation. MLP is Multi-Layer-Perceptron.}
\end{figure*}

\Fig{VidConvPipeline} illustrates the overall architect of VidConv model. The VidConv takes as input a clip $X \in \mathbb{R}^{L \times H \times W \times 3}$ consisting of $L$ RGB frames of size $H \times W$ sampled from the original video. The network has a stem layer and 4 stages, after which the feature is down-sampled by a half. The network's stages can then be split into two parts. 

The early-net includes the stem layer, stage \nth{1} and \nth{2} as in the original 2D ConvNeXt \cite{liu2022convnet}. It processes a batch of $L$ frames as regular 2D images, and outputs $L$ features of size $H/8 \times W/8 \times C$.

However, before passing the feature into stage \nth{3} and \nth{4} of the later-net, we first rearrange the feature of $L=hw$ frames into a grid of $h \times w$, forming a ``collage feature" of size $hH/8 \times wW/8 \times C_1$. For example, for $L=9$, we arrange the features into $h=3$ vertically and $w=3$ horizontally. The later-net then processes the collage feature as it is extracted from a single 2D image, and output a tensor of shape $hH/32 \times wW/32 \times C_2$. 

Intuitively, this design casts the problem from recognizing a video of $L$ frames into classifying a collage image with $h\times w$ higher resolutions. The motivation is that modern ConvNet and ViT often use large kernel size, \eg $7 \times 7$, and more number blocks in stage \nth{3} and \nth{4}, \eg ConvNeXt-small models has 27 blocks in stage \nth{3}. Hence, flattening the 3D feature of a video into 2D form is the simplest way to capture temporal structure by exploiting the large receptive field of networks.
 
In addition, we introduce the VidConv block in the later-net to increase the temporal connection between the frames. As illustrated in \fig{VidConvBlock}, the VidConv block is a minimal modification of the ConvNeXt block, in which we add a depth-wise convolution of kernel size $(h,w)$ and dilation $(H,W)$ without padding. 

Concretely, the VidConv block takes an input feature of size  $hH \times wW \times C$. Then, the depth-wise convolution $7\times 7$ extracts the spatial features $S \in hH \times wW \times C$, and the dilated (also depth-wise) convolution extracts the temporal feature $T \in H \times W \times C$. The spatial-temporal fusion is then obtained by adding $T$ and $S$:
\begin{equation}\label{eq:fusion}
F = S + \alpha * [T]_{(\times h, \times w)}
\end{equation} 
where $[\cdot]_{(\times h, \times w)}$ denotes repeating the tensor $h$ times horizontally and $w$ vertically, $\alpha \in \Re^C $ is the learnable vector to balance between temporal and spatial branch, and $*$ is the element-wise product. This design allows sharing the temporal features $T$ for all the frames. Feature $F$ is then passed to Norm Layer and MLP layers. 

In fact, there can be many variants for this design, such as using non-shared weight convolution, dynamics convolution \cite{chen2020dynamic}, deformable convolution \cite{dai2017deformable} or attention blocks \cite{vaswani2017attention}. Using depth-wise dilated convolution is one of the most efficient and simplest options. The design is inspired from our hypothesis that MLP can be the main factor contributing to success of ViT, since MLP constitutes more then two-thirds of network's parameters and it is the only unit having Nonlinear activation. For the same reason, PoolFormer \cite{yu2021metaformer} only uses Pooling operator to perform spatial connection, or ConvMixer \cite{trockman2022patches} simply uses depth-wise convolution. We follows this minimalist design.
    
Finally, the feature extracted from later-net is fed to a neck, which is implemented as a simple convolution of kernel size $(h,w)$, dilation $(H_2,W_2)$ without padding to reduce the spatial size and aggregate the temporal features before global pooling.      
   
\subsection{Architecture Variants and Pre-trained Weights}
\begin{figure*}[t]
	\centering
	\includegraphics[width=0.8\textwidth]{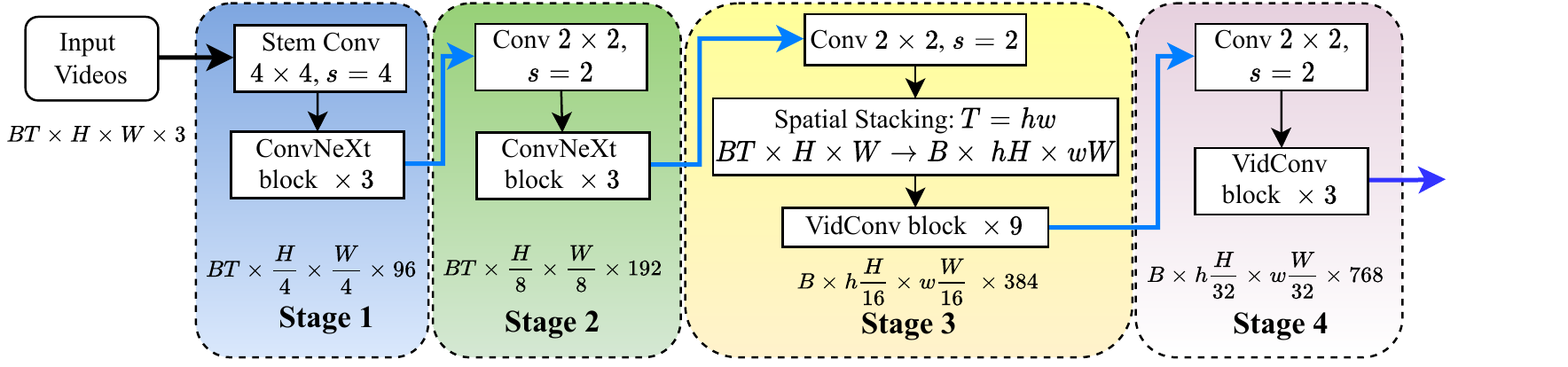}
	\caption{Block diagram of VidConv-tiny, referred as ViC-T. $B$ denotes for number of clips, and $T$ for number of sampled frames in a clip, $(H,W)$ is the image size. $k \times k$ denotes the kernel size, $s$ is the stride of convolution, and block $\times n$ means stacking $n$ blocks in sequence.}
	\label{fig:VidConvTiny}
\end{figure*}

In this work, we focus on small and medium size models. Our VidConv Model inherits the configuration of ConvNeXt models \cite{liu2022convnet}, and \tab{ModelConfig} summarizes the architecture of these model variants. \Fig{VidConvTiny} illustrates the block configuration of VidConv-Tiny.

\begin{table}[h]
	\caption{VidConv Model Configuration.}
	
	\label{tab:ModelConfig}
	\centering
	\begin{adjustbox}{max width=\columnwidth}
		\small{
	\begin{tabular}{|l|c|c|c|}
		\hline
		Model         & Channel of each stage & Number of Blocks & $C_3$ dim \\ \hline
		Tiny (ViC-T)  & $(96,192,384,768)$      & $(3,3,9,3)$        & $2304$     \\ \hline
		Small (ViC-S) & $(96,192,384,768)$      & $(3,3,27,3)$        & $2304$     \\ \hline
		Base (ViC-B)  & $(128,256,512,1024)$    & $(3,3,27,3)$       & $2304$     \\ \hline
	\end{tabular}}
	\end{adjustbox}
\end{table}

Finally, keeping the configuration similar to the original ConvNeXt allows using their strong pre-trained ImageNet for network initialization. Hence, we only need to initialize the dilated convolutions in VidConv blocks from random. 

\section{Experiments}
\subsection{Datasets} We evaluate VidConv on 3 benchmark action recognition datasets: Kinetics-400 (K400) \cite{carreira2017quo}, Kinetics-600 (K600) \cite{carreira2018short} and Something-Something V2 (SSv2) \cite{goyal2017something}. Kinetics datasets consist of 10-second videos sampled at 25fps from YouTube. K400 has about 240k training and 20k validation videos in 400 human action categories. K600 is the extension of K400, and consists of 370k training videos and 28.3k validation videos in 600 human action categories. SSv2 contains 220k videos sampled at 12fps with duration ranging from 2 to 6 seconds. The dataset is split to 169k training and 24.7k validation videos over 174 fine-grain action categories. Follow the prior works, we report the accuracy Top-1 and Top-5 on the validation set for all methods. \\

\subsection{Implementation Details} \label{sec:Implement}Our code is implemented using MMAction2 framework \cite{2020mmaction2}. We train the models with AdamW optimizer \cite{loshchilov2018fixing} with a batch size of 64. We use a cosine decay learning rate scheduler and linear warm-up in the first epoch. The initial learning rate is 1e-3, and decays to the minimal value 5e-6. To regularize the model, we use the stochastic depth with drop path rate 0.25 for ViC-T, 0.4 for ViC-S and 0.5 for ViC-B models. We apply standard image augmentation including left-right flip and multi-scale cropping. Because the temporal dilated convolution is initialized randomly, we set the initial value for $\alpha$ in \eqn{fusion} to 1e-2. In addition, follow \cite{liu2021video}, since the head and the final dilated convolution are initialized randomly, we multiply the learning rate of backbone by a factor $l_b$ to stabilize the training.

For K400 and K600, we use the models pretrained on ImageNet-1k for ViC-T and ViC-S, and models pretrained on ImageNet-21k for ViC-B for initialization. We train the models for 24 epochs, and set $l_b$ to 0.25 for ViC-T, and 0.1 for ViC-S and ViC-B models.  Unless otherwise mentioned, for all model variants, we sample a clip of 9 frames from each full length video using a random temporal stride in range of 8 to 12, and spatial size of $224 \times 224$. For inference, we randomly sample 4 clips from a full video, each clip has 9 frames using a fixed temporal stride of 12. The final score is computed as the average score over all the views. 

For SSv2, follow the previous works, we use the models pretrained on K400 for initialization. However, we found that the models are prompt to be overfit. Hence, we only train the tiny and small models for 20 epochs with early-stopping, and set $l_b$ to 0.25 for both ViC-T and ViC-S. We also apply strong augmentation with RandAugment \cite{cubuk2020randaugment}, and drop-out after the global pooling with ratio 0.65 for ViC-T and 0.8 for ViC-S. Unless otherwise mentioned, for all model variants, we sample a clip of 9 frames from each full length video using a random temporal stride $s_T$ in range of 5 to 7, and spatial size of $224 \times 224$. For inference, we randomly sample 2 clips from a full video, each clip has 9 frames using a fixed temporal stride of 6. For any videos that don't have enough frames, we select the maximum temporal stride that can fit entire the videos. Concretely, $s_T=\min(5,(N-1)//8)$ where $N$ is the number of frames. The final score is computed as the average score of 2 clips with 3 random crops.

\subsection{Comparison to state-of-the-art}
\subsubsection{Notation} In this section, ``Views" indicates number temporal clips $\times$ number spatial crop. For example, the notation of $4\times 3$ views indicates a video is split into 4 clips with different start indices, and frames in each clip are cropped into left, center and right portions, yielding total 12 views. Frames (number), FLOPs (Giga $10^9$) and Latency (ms) are reported for a single view, and the \textbf{corresponding total amounts are multiplied by the number of views}. The latency is reported on a single GPU Nvidia-2080Ti, and the parameters are in Mega ($10^6$). Except the TimeSformer-HR uses image size of $448\times 448$, other models use image size of $224 \times 224$ or $256 \times 256$. IN-1K and IN-21K denotes pretrained on ImageNet-1k and ImageNet-21k \cite{deng2009imagenet} respectively. \textbf{Bolt} and \textbf{\textit{Bold-Italic}} indicates the highest and second highest score ($\pm 0.1$) in each group. 
\subsubsection{Kinetics-400} \Tab{SOTA-K400} compares our solution on Kinetics-400 to the state-of-the-art (SOTA) Convolution and Transformer based methods in the first and second part of the table, respectively. Convolution models typically require much longer training epochs (100-300 epochs), and more frames and views (1$\times$ 10 to $10\times3$) for equal performance. For transformer-based models, although having less number of parameters, their FLOP and latency are typically higher. TimeSformer-L, Swin-S, Swin-B, Uniform-S achieve accuracy $80.7 (\pm0.1)$ Top-1 and 94.7 Top-5, but trained and inference with more frames per views, \eg TimeSformer-L uses 96 frames per clip or Swin models are inferred with 12 clips of 32 frames. Our base model ViC-B also achieves 80.2 Top-1 and 94.4 Top-5 accuracy using only 4 clips of 9 frames input, closely match the performance of TimeSformer and Swin models. Note that, inference speed of ViC models is significantly faster than ViT-based models because during performing sliding-window feature extraction, we only need to run the early-net for a new coming frame, while the low-level feature of the previous frames were already extracted and can be drawn from a small buffer. This can be achieved because the early-net process each frame independently.

\begin{table*}[!t]
	\caption{Comparison to state-of-the-art on Kinetics-400. }
	\label{tab:SOTA-K400}
	\centering
	\begin{adjustbox}{max width=\textwidth}
		\small{
	\begin{tabular}{|l||c|c||c|c||c|c|c|c|c|}
		\hline
		Method         & Pretrain & \begin{tabular}[c]{@{}c@{}}Training \\ Epochs\end{tabular} & Top-1 & Top-5 & Views & \begin{tabular}[c]{@{}c@{}}Frames\\ /view\end{tabular} & \begin{tabular}[c]{@{}c@{}}FLOPs\\ /view \end{tabular} & \begin{tabular}[c]{@{}c@{}}Latency \\ /view \end{tabular} & \begin{tabular}[c]{@{}c@{}}Params \\ \end{tabular} \\ \hline
		TSN-R50 \cite{wang2016temporal}      & IN-1K    & 100                                                        & 70.6  & 89.2  & 1$\times$10  & 25                                                     & 103                                                       & 31.0                                                          & 24.3                                                 \\ 
		TANet-R50 \cite{liu2021tam}       & IN-1K    & 100                                                        & 76.3  & 92.6  & 1$\times$3   & 8                                                      & 43.1                                                      & 13.8                                                          & 25.6                                                 \\ 
		I3D-R50 \cite{carreira2017quo}        & IN-1K    & 100                                                        & 72.68 & 90.78 & 10$\times$3  & 32                                                     & 43.5                                                      & 25.7                                                          & 28.0                                                 \\ 
		SlowFast-8x8 \cite{feichtenhofer2019slowfast}  & -        & 196                                                        & 77    & 92.6  & 10$\times$3  & 32                                                     & 65.8                                                      & 38.1                                                          & 34.5                                                 \\ 
		SlowFast-16x8\cite{feichtenhofer2019slowfast}  & -      & 256                                                        & 78.9  & 93.5  & 10$\times$3  & 32                                                     & 213                                                      & N/A                                                          & N/A                                                 \\
		X3D-M \cite{feichtenhofer2020x3d}         & -        & 300                                                        & 76    & 92.3  & 10$\times$3  & 16                                                     & 6.2                                                       & 12.1                                                         & 3.8                                                  \\ 
		X3D-XL \cite{feichtenhofer2020x3d}        & -        & 300                                                        & \textbf{79.1}  & 93.9  & 10$\times$3  & 16                                                     & 48.4                                                      &            N/A                                               & 11.0                                                  \\ 
		TSM-R50 \cite{lin2019tsm}       & IN-1K    & 100                                                        & 74.7  & -     & 10$\times$3  & 16                                                     & 65.9                                                      & 22.9                                                          & 24.3                                                 \\ \hline
		TimeSformer \cite{gberta_2021_ICML}   & IN-1K    & 15                                                         & 78    & 93.7  & 1$\times$3   & 8                                                      & 100.9                                                     & 50.5                                                          & 121.4                                                \\ 
		TimeSformer-HR \cite{gberta_2021_ICML}& IN-1K    & 15                                                         & \textit{79.7}  & 94.4  & 1$\times$3   & 16                                                     & 807.2                                                     & 403.2                                                         & 121.4                                                \\ 
		TimeSformer-L \cite{gberta_2021_ICML} & IN-21K   & 15                                                         & \textit{\textbf{80.7}}  & 94.7  & 1$\times$3   & 96                                                     & 1210.8                                                    & 515.4                                                         & 121.4                                                \\ 
		Swin-T \cite{liu2021video}         & IN-1K    & 30                                                         & 78.8  & 93.6  & 4$\times$3   & 32                                                     & 88                                                        & 57.7                                                          & 28.2                                                 \\ 
		Swin-S \cite{liu2021video}         & IN-1K    & 30                                                         & \textit{\textbf{80.6}}  & 94.5  & 4$\times$3   & 32                                                     & 166                                                       & 93.5                                                         & 49.8                                                 \\ 
		Swin-B \cite{liu2021video}         & IN-1K    & 30                                                         & \textit{\textbf{80.6}}  & 94.6  & 4$\times$3   & 32                                                     & 282                                                       & 127.9                                                         & 88.1                                                 \\ 
		MViT-B 16$\times$4 \cite{fan2021multiscale}    & -    & 200                                                         & 78.4  & 93.5  & 5$\times$1   & 16                                                     & 70.5                                                       & 28.7                                                         & 36.6                                                 \\ 
		MViT-B 32$\times$4 \cite{fan2021multiscale}    & -    & 200                                                         & 80.2  & 94.4  & 5$\times$1   & 32                                                     & 170                                                       & 86.6                                                        & 36.6                                                 \\ 
		Uniformer-S \cite{li2022uniformer}    & IN-1K     & 110          & \textit{\textbf{80.8}}  & 94.7  & 4$\times$1   & 16                                                     & 167                                                       & ??                                                        & 21.4                                                 \\ 
		Uniformer-B \cite{li2022uniformer}   & IN-1K     & 110           & \textbf{82.0}  & 95.1  & 4$\times$1   & 16                                                     & 389                                                       & ??                                                        & 49.4                                                 \\ \hline \hline
		ViC-T [Ours]         & IN-1K    & 24                                                         & 75.4  & 92.2  & 4$\times$1   & 9                                                      & 40.9                                                      & 11.4                                                          & 44.7                                                 \\ 
		ViC-S [Ours]        & IN-21K    & 24                                                         & 78.2  & 93.5  & 4$\times$1   & 9                                                      & 79.0                                                      & 22.4                                                          & 66.4                                                 \\ 
		ViC-S [Ours]        & IN-21K    & 24                                                         & 79.5  & 94.0  & 4$\times$1   & 16                                                      &140.4                                                       &                                                           &   78.9                                               \\
		ViC-B [Ours]        & IN-21K   & 24                                                         & \textbf{\textit{80.2}}  & 94.4  & 4$\times$1   & 9                                                      & 139.2                                                     & 33.8                                                          & 107.4                                                \\ 
		ViC-B [Ours]        & IN-21K   & 24                                                         & \textbf{80.5}  & 94.5  & 4$\times$3   & 9                                                      & 139.2                                                     & 33.8                                                          & 107.4                                                \\ \hline
	\end{tabular}}
	\end{adjustbox}
	
%
	\vspace{8mm}
	\caption{Comparison to state-of-the-art on Kinetics-600. }
	\label{tab:SOTA-K600}
	\centering
	\begin{adjustbox}{max width=\textwidth}
		\small{
			\begin{tabular}{|l|c|c|cc|cc|c|c|c|c|}
				\hline
				\multirow{2}{*}{Method} & \multirow{2}{*}{Pretrain} & \multirow{2}{*}{\begin{tabular}[c]{@{}c@{}}Training \\ Epochs\end{tabular}} & \multicolumn{2}{c|}{\begin{tabular}[c]{@{}c@{}}Official \\ Report\end{tabular}} & \multicolumn{2}{c|}{\begin{tabular}[c]{@{}c@{}}Evaluate on \\ CVDF-K600\end{tabular}} & \multirow{2}{*}{Views} & \multirow{2}{*}{\begin{tabular}[c]{@{}c@{}}Frames\\ /view\end{tabular}} & \multirow{2}{*}{\begin{tabular}[c]{@{}c@{}}FLOPs\\ /view\end{tabular}} & \multirow{2}{*}{Params} \\ \cline{4-7}
				&                           &                                                                             & \multicolumn{1}{c|}{Top-1}                        & Top 5                       & \multicolumn{1}{c|}{Top-1}                           & Top-5                          &                        &                                                                         &                                                                        &                         \\ \hline
				SlowFast 16×8 +NL  \cite{feichtenhofer2019slowfast}     & -                         & 196                                                                         & \multicolumn{1}{c|}{\textbf{81.8}}                         & 95.1                        & \multicolumn{1}{c|}{-}                               & -                              & 10$\times$3                   & 16                                                                      & 234                                                                    & 59.9                    \\ 
				X3D-M  \cite{feichtenhofer2020x3d}                 & -                         & 256                                                                         & \multicolumn{1}{c|}{78.8}                         & 94.5                        & \multicolumn{1}{c|}{-}                               & -                              & 10$\times$3                    & 16                                                                      & 6.2                                                                    & 3.8                     \\ 
				X3D-XL  \cite{feichtenhofer2020x3d}                & -                         & 512                                                                         & \multicolumn{1}{c|}{\textbf{81.9}}                         & 95.5                        & \multicolumn{1}{c|}{-}                               & -                              & 10$\times$3                    & -                                                                       & 48.4                                                                   & 11                      \\ \hline
				TimeSformer  \cite{gberta_2021_ICML}           & IN-21K                    & 15                                                                          & \multicolumn{1}{c|}{79.1}                         & 94.4                        & \multicolumn{1}{c|}{84.6}                           & 96                             & 1$\times$3                     & 8                                                                       & 196                                                                    & 121.4                   \\ 
				TimeSformer-HR \cite{gberta_2021_ICML}         & IN-21K                    & 15                                                                          & \multicolumn{1}{c|}{81.8}                         & 95.8                        & \multicolumn{1}{c|}{\textbf{88}}                              & 97.2                           & 1$\times$3                     & 16                                                                      & 1703                                                                   & 121.4                   \\ 
				TimeSformer-L \cite{gberta_2021_ICML}          & IN-21K                    & 15                                                                          & \multicolumn{1}{c|}{82.2}                         & 95.6                        & \multicolumn{1}{c|}{\textbf{87.9}}                            & 97                             & 1$\times$3                     & 96                                                                      & 2308                                                                   & 121.4                   \\ 
				Swin-B  \cite{liu2021video}                & IN-21K                    & 30                                                                          & \multicolumn{1}{c|}{\textbf{84}}                           & 96.5                        & \multicolumn{1}{c|}{87.2}                            & 97.3                           & 4$\times$3                     & 32                                                                      & 282                                                                    & 88.1                    \\ 
				MViT-B, 16x4  \cite{fan2021multiscale}          & -                         & 200                                                                         & \multicolumn{1}{c|}{82.1}                         & 95.7                        & \multicolumn{1}{c|}{-}                               & -                              & 5$\times$1                    & 16                                                                      & 70.5                                                                   & 36.8                    \\ 
				MViT-B, 32x3  \cite{fan2021multiscale}          & -                         & 200                                                                         & \multicolumn{1}{c|}{83.4}                         & 96.3                        & \multicolumn{1}{c|}{-}                                & -                               & 5$\times$1                    & 32                                                                      & 170                                                                    & 36.8                    \\ 
				MViT-B-24, 32x3 \cite{fan2021multiscale}        & -                         & 200                                                                         & \multicolumn{1}{c|}{\textbf{83.8}}                         & 96.3                        & \multicolumn{1}{c|}{-}                               & -                              & 5$\times$1                    & 32                                                                      & 236                                                                    & 52.9                    \\ 
				UniFormer-B             & IN-1K                     & 200                              & \multicolumn{1}{c|}{\textbf{84}}               & 96.4             & \multicolumn{1}{c|}{83.3}               & 93.4               & 4x1                    & 16                           & 389                         & 49.4                    \\
				UniFormer-B             & IN-1K                     & 200                              & \multicolumn{1}{c|}{\textbf{83.8}}             & 96.7             & \multicolumn{1}{c|}{84.1}                & 93.6               & 4x1                    & 32                           & 1036                        & 49.4                    \\
				\hline \hline
				ViC-B [Ours]                   &        IN-21K                   &           24                                                                  & \multicolumn{1}{c|}{-}                            & -                           & \multicolumn{1}{c|}{86.1}                            & 96.2                           & 4$\times$1                    & 9                                                                       & 139.2                                                                  & 107.4                  \\ \hline
			\end{tabular}
		}
	\end{adjustbox}
\end{table*}

\subsubsection{Kinetics-600} We select model ViC-B for comparison with other state-of-the-art methods. The experiment setup and training parameters are similar to K400 presented in \sec{Implement}, except that we reduce the stochastic depth dropout ratio to 0.25 for the backbone, and the drop-out ratio to 0.2 for the head. Since the dataset K600 is no longer provided in the official website, we downloaded it from the CVDFoundation \cite{cvdf2022kinetics}. We denote this version as CVDF-K600 due to a slight mismatch with the original release. Concretely, the validation set of CVDF-K600 has about 29.7k videos, but only 17k videos are in the official version. Therefore, we include the results on two validation set versions if the checkpoints of other methods are available. 

As shown in \tab{SOTA-K600}, our model ViC-B achieves 86.1 Top-1 and 96.2 Top-5 accuracy, and performs better than TimeSformer and Uniformer-B on the CVDF-K600 version. Note that, most of other methods get better score when re-evaluating than their official report, except for the Uniformer that slightly change. The TimeSformer-HR,TimeSformer-L and Swin-B gets the highest score (\eg 88 Top-1) but with much higher FLOPs and more views.


\subsubsection{Somthing-Something V2} \Tab{SOTA-SSv2} compares our approach with the SOTA methods on SS-v2. Compare to Convolution based methods, our ViC-S model can closely match the performance of STM \cite{jiang2019stm}, MSNet \cite{kwon2020motionsqueeze} and TEA \cite{li2020tea}, which have more intricate modules designed to capture the long-range relationship. TEA \cite{li2020tea} model gets the highest Top-1 accuracy (65.1) in this group but is tested with 5x more views ($10\times3$). Compared to ViT based methods, our ViC-S model with 9-frames input has top-1 64.4, higher accuracy than TimeSformer \cite{gberta_2021_ICML} with significantly less FLOPs. When using 16 frames input, our ViC-S model achieves Top-1 accuracy 65.1, 2\% lower than MViT-B\cite{fan2021multiscale} model (67.1). Swin-B \cite{liu2021video} and Uniformer \cite{li2022uniformer} get the highest accuracy in this group, but require more frames and FLOPs.
    
\begin{table*}
	\caption{Comparison to state-of-the-art on Something-Something V2. }
	\label{tab:SOTA-SSv2}
	\centering
	\begin{adjustbox}{max width=\textwidth}
		\small{
		\begin{tabular}{|l||c|c||c|c||c|c|c|c|}
			\hline
			Method            & Pretrain & \begin{tabular}[c]{@{}c@{}}Training \\ Epochs\end{tabular} & Top-1 & Top 5 & Views & \begin{tabular}[c]{@{}c@{}}Frames\\ /view\end{tabular} & \begin{tabular}[c]{@{}c@{}}FLOPs\\ /view\end{tabular} & Params \\ \hline
			SlowFast R101 8$\times$8 \cite{feichtenhofer2019slowfast} & K-400    & 22                                                         & 63.1  & 87.6  & 1$\times$3   & 64                                                     & 106                                                   & 53.3   \\ 
			TSM-R50 \cite{lin2019tsm}          & K-400    & 50                                                         & 63.3  & 88.2  & 2$\times$3   & 16                                                     & 62                                                    & 42.9   \\ 
			STM-R50 \cite{jiang2019stm}          & IN-1K    & 50                                                         & 64.2  & 89.8  & 10$\times$3  & 16                                                     & 66.5                                                  & 24.0     \\
			MSNet-R50 \cite{kwon2020motionsqueeze}     & IN-1K    & 40                                                         & 64.7  & 89.4  & 1$\times$1   & 16                                                     & 67                                                    & 24.6   \\ 
			TANet-R50 \cite{liu2021tam}    & IN-1K    & 50                                                         & 64.6  & 89.5  & 2$\times$3   & 16                                                     & 66                                                    & -   \\
			TEA-R50 \cite{li2020tea}          & IN-21K   & 50                                                         & \textbf{65.1}  & \textbf{89.9}  & \textbf{10$\times$3}  & 16                                                     & 70                                                    & -      \\ \hline
			TimeSformer  \cite{gberta_2021_ICML}     & IN-21K   & 15                                                         & 59.5  & -     & 1$\times$3   & 8                                                      & 101                                                   & 121.4  \\ 
			TimeSformer-HR \cite{gberta_2021_ICML}   & IN-21K   & 15                                                         & 62.2  & -     & 1$\times$3   & 16                                                     & 807.2                                                 & 121.4  \\ 
			TimeSformer-L \cite{gberta_2021_ICML}    & IN-21K   & 15                                                         & 62.4  & -     & 1$\times$3   & 64                                                     & 1210.9                                                & 121.4  \\ 
			Swin-B \cite{liu2021video}           & K-400    & 60                                                         & \textit{\textbf{69.6}}  & \textbf{92.7}  & 1$\times$3   & 32                                                     & 282                                                   & 88.1   \\ 
			MViT-B 16$\times$4 \cite{fan2021multiscale}      & K-400    & 100                                                     & 64.7  & 89.2  & 1$\times$3   & 16                                                     & 70.5                                                & 36.6   \\
			MViT-B 32$\times$3 \cite{fan2021multiscale}      & K-400    & 100                                                     & 67.1  & 90.8  & 1$\times$3   & 32                                                     & 170                                                & 36.6   \\ 
			Uniformer-S \cite{li2022uniformer}      & K-400    & 200                                                     & 67.7  &  88.5 & 1$\times$3   & 16                                                     & 125                                                & 21.4   \\
			Uniformer-B \cite{li2022uniformer}      & K-400    & 200                                                     & \textbf{70.4}  &  \textbf{92.8} & 1$\times$3   & 16                                                     & 290                                                & 49.4   \\ \hline \hline
			ViC-T [Ours]       & K-400    & 20                                                         & 61.3  & 86.9  & 1$\times$3 & 9                                                      & 40.9                                                 & 44.2  \\ 
			ViC-S [Ours]       & K-400    & 20                                                         & 64.4  & 88.8  & 1$\times$3  & 9                                                      & 79.0                                                 & 65.9  \\ 
			ViC-S [Ours]       & K-400    & 20                                                         & \textbf{65.1}  & \textbf{89.6}  & 1$\times$3  & 16                                                      & 140.4                                             & 78.4  \\ \hline
		\end{tabular}}
	\end{adjustbox}
\end{table*}

\subsubsection{Training Efficiency} One important criteria to select models for production is the training efficiency. Although longer training with strong data augmentation and careful tuning can sometimes improve the performance 1-2\% on benchmark dataset, it is hardly considered in industry due to the cost expense and delaying time.
We therefore compare the real training time between the models using the same server with configuration of 4 GPUs Quadro RTX 8000 (48G), 256GB RAM and Intel(R) Xeon(R) Silver 4210R CPU @ 2.40GHz (20 cores). For all other methods, the results are reported by using their official released code. The training time for dataset K400 is shown in \fig{AccTrainingTime}, and \tab{TrainingTime} shows the training time and memory consumed for K600 dataset. From \tab{TrainingTime}, we see that for the same amount training epoch, the more frames in a clip, the longer training time, and data loading time becomes the main bottleneck. Our method gets the best compromise between the consumed training resource and the accuracy. It only needs 46 GPU hours (2 days) to achieve 86.1 top-1 accuracy, while Swin-B would need 216 GPU hours (9 days) and TimeSformer and UniFormer need 626 hours (26 days) and 1824 hours (76 days) respectively.   
\begin{table*}[!h]
	\caption{Compare the consuming training resource of state-of-the-art methods.   }
	\label{tab:TrainingTime}
	\centering
	\begin{adjustbox}{max width=\textwidth}
		\small{
			\begin{tabular}{|l|c|c|c|c|c|c|c|}
				\hline
				Method            & \begin{tabular}[c]{@{}c@{}}Training \\ Epochs\end{tabular} & \begin{tabular}[c]{@{}c@{}}Batch \\ size\end{tabular} & \begin{tabular}[c]{@{}c@{}}Frames\\ /clip\end{tabular} & \begin{tabular}[c]{@{}c@{}}Image\\ Size\end{tabular} & \begin{tabular}[c]{@{}c@{}}GPU Memory\\ (GB/card)\end{tabular} & \begin{tabular}[c]{@{}c@{}}Training time \\ (GPU hours)\end{tabular} & Top-1 \\ \hline
				TimeSformer       & 15                                                         & 64                                                    & 8                                                      & 224                                                  & 10.3                                                           & 101                                                              & 84.62 \\ 
				TimeSformer-HR    & 15                                                         & 24                                                    & 16                                                     & 448                                                  & 25.4                                                           & 518                                                              & 88    \\ 
				TimeSformer-L     & 15                                                         & 24                                                    & 96                                                     & 224                                                  & 26.3                                                           & 626                                                              & 87.9  \\ 
				Swin-B            & 30                                                         & 64                                                    & 32                                                     & 224                                                  & 21.4                                                           & 216                                                              & 87.2  \\ 
				UniFormer-B, 16x4 & 110                                                        & 16                                                    & 16                                                     & 224                                                  & 25.3                                                           & 816                                                              & 83.3  \\ 
				UniFormer-B, 32x4 & 110                                                        & 8                                                     & 32                                                     & 224                                                  & 41.6                                                           & 1824                                                             & 84.1  \\ \hline \hline
				ViC-B [Ours]   & 24                                                         & 64                                                    & 9                                                      & 224                                                  & 24.2                                                           & \textbf{46}                                                               & 86.1  \\ \hline
			\end{tabular}
		}
	\end{adjustbox}
\end{table*}
\subsection{Ablation Study}
\subsubsection{Default Model and Training Setting}
In this section, we investigate different factors affecting the performance of ViC. We use ViC-Tiny pretrained in ImageNet-1k and 9 frames per input clip as the default model. Unless otherwise mentioned, we drop the neck, which is the final dilated convolution in \fig{VidConvPipeline} in order to isolate its effect from the backbone. We train the model for 24 epochs on K400 and 20 epochs on Sth-Sth-V2 using standard augmentation, including left-right flip and multi-scale cropping with image size $224 \times 224$. The results are reported with 4$\times$ 1 views on Kinetics-400, and $1\times3$ views on Sth-Sth V2 dataset. 

\subsubsection{Temporal branch and Spatial Stacking} We examine independent effect of the spatial stacking and the temporal branch, which is realized by the simple dilated-convolution in \fig{VidConvBlock}. The result is reported in \Tab{TempEffect}. The \nth{1} row show the simplest baseline, where we just use the 2D ConvNeXt model on each frame separately, then perform global pooling to extract feature. 

\begin{table}[h]
	\caption{Effect of Temporal branch and spatial stacking.}
	\label{tab:TempEffect}
	\centering
	\begin{adjustbox}{max width=\columnwidth}
	\small{
	\begin{tabular}{|c|c|cc|cc|}
		\hline
		\multirow{2}{*}{\begin{tabular}[c]{@{}c@{}}Temporal \\ Branch\end{tabular}} & \multirow{2}{*}{\begin{tabular}[c]{@{}c@{}}Spatial \\ Stacking\end{tabular}} & \multicolumn{2}{c|}{Kinetics-400}                  & \multicolumn{2}{c|}{Sth-Sth V2}                    \\ \cline{3-6} 
		&                                                                              & \multicolumn{1}{c|}{Top-1} & Top-5                 & \multicolumn{1}{c|}{Top-1} & Top-5                 \\ \hline
		\multicolumn{1}{|c|}{None}                                                  & \multicolumn{1}{c|}{None}                                                    & \multicolumn{1}{c|}{72.2}      & \multicolumn{1}{c|}{89.6} & \multicolumn{1}{c|}{39.8}      & \multicolumn{1}{c|}{72.4} \\ \hline
		None                                                                        & $3 \times 3$                                                                          & \multicolumn{1}{c|}{73.8}  & 90.6                  & \multicolumn{1}{c|}{53.2}  & 80.8                  \\ \hline
		Yes                                                                         & None                                                                         & \multicolumn{1}{c|}{73.2}  & 90.2                  & \multicolumn{1}{c|}{58.2}  & 84.6                      \\ \hline
		Yes                                                                         & $3 \times 3$                                                                          & \multicolumn{1}{c|}{74.5} & 91.4                  & \multicolumn{1}{c|}{60.4}  & 86.7                  \\ \hline
	\end{tabular}}
	\end{adjustbox}	
\end{table}
In the \nth{2} row, we apply feature stacking strategy but still using the original ConvNeXt model. In the \nth{3} row, we use VidConv block in later-net without stacking the feature along spatial axes. Concretely, only the dilated-convolution is applied on stacked feature, while the depth-wise conv $7\times7$ operates on each frame independently. This is equivalent to the design of (2+1)D convolution, where we apply two depth-wise 3D convolutions $1\times7\times7$ and $T\times1 \times1$ to the input $C\times T\times H \times W$. Finally, the last row shows our ViC-T using both spatial stacking and VidConv block. We see that both strategy can improve the performance compared to the baseline, and their combination can further improve the results significantly. The temporal branch is especially useful on the SS-v2 dataset. This is because the action categories in SS-v2 are more temporal dependent than that of Kinetics-K400. 

\subsubsection{Spatial Resolution} \Tab{SpatialEffect} compares the accuracy when increasing the number of frames and spatial stacking resolution. 

\begin{table}[t]
	\caption{Effect of spatial stacking resolution.}
	\label{tab:SpatialEffect}
	\centering
	\small{
		\begin{tabular}{|c|cc|cc|}
			\hline
			\multirow{2}{*}{\begin{tabular}[c]{@{}c@{}}Spatial \\ Stacking\end{tabular}} & \multicolumn{2}{c|}{Kinetics-400}  & \multicolumn{2}{c|}{Sth-Sth V2}    \\ \cline{2-5} 
			& \multicolumn{1}{c|}{Top-1} & Top-5 & \multicolumn{1}{c|}{Top-1} & Top-5 \\ \hline
			None                                                                         & \multicolumn{1}{c|}{73.2}      &   90.2    & \multicolumn{1}{c|}{58.2}  & 84.6  \\ \hline
			$2 \times 2$                                                                          & \multicolumn{1}{c|}{73.9} & 90.7 & \multicolumn{1}{c|}{56.2}   & 83.5     \\ \hline
			$3 \times 3$                                                                          & \multicolumn{1}{c|}{74.5} & 91.4  & \multicolumn{1}{c|}{60.4}  & 86.7  \\ \hline
			$4 \times 4$                                                                          & \multicolumn{1}{c|}{75.5} & 92.0  & \multicolumn{1}{c|}{62.3}  & 87.7   \\ \hline
	\end{tabular}}
\end{table}
The \nth{1} row is the same as the \nth{3} row of \Tab{TempEffect}, where we use 9 frames per clip with the temporal branch but without spatial stacking. We denote this case as 1$\times$1 for later reference. 2$\times$2 is applied when we use only 4 frames and 2 clips (\eg total 8 frames) during training and testing. 3$\times$3 and 4$\times$4 denotes the default configuration with 9 frames and 16 frames per clip, respectively. We see that for the same number of frames, computation and parameters, 3$\times$3 configuration improves the result relatively to 1$\times$1. The performance of 2$\times$2 is better than 1$\times$1 and lower than 3$\times$3 on K600 dataset as expected. However, on SSv2 dataset, the case 1$\times$1 is surprisingly better than 2$\times$2. The reason is that although 1$\times$1 case does not apply spatial stacking, it extract the temporal feature from 9 frames, while 2$\times$2 cases only use 4 frames. Since on SSv2 dataset are more temporal dependent, the temporal branch is dominant in this case. The 4$\times$4 configuration consistently achieves the highest score on both datasets, proving the importance of number of frames input.   

\subsubsection{Stacking Level} In \tab{StackingLevel}, we investigate the optimal stage to perform spatial stacking. Stacking after \nth{4} stage is equivalent to the baseline 2D ConvNeXt (\nth{1} row of \tab{TempEffect}). We see that the earlier we perform stacking feature, the better temporal information is extracted. Thus, the higher accuracy but also more computation and longer latency. 

\begin{table}[t]
	\caption{Effect of spatial stacking at different stages. Stage \nth{2} is the default as shown in \fig{VidConvTiny}. }
	\label{tab:StackingLevel}
	\centering
	\setlength{\tabcolsep}{4pt}
	\begin{adjustbox}{max width=\columnwidth}
	\small{
	\begin{tabular}{|c|cc|cc|c|c|c|}
		\hline
		\multirow{2}{*}{Stages} & \multicolumn{2}{c|}{Kinetics-400}  & \multicolumn{2}{c|}{Sth-Sth V2}    & \multirow{2}{*}{\begin{tabular}[c]{@{}c@{}}FLOPs\\ /view\end{tabular}} & \multirow{2}{*}{\begin{tabular}[c]{@{}c@{}}Latency\\ /view\end{tabular}} & \multirow{2}{*}{Params} \\ \cline{2-5}
		& \multicolumn{1}{c|}{Top-1} & Top-5 & \multicolumn{1}{c|}{Top-1} & Top-5 &                                                                        &                                                                          &                        \\ \hline
		\nth{1}                                                                                & \multicolumn{1}{c|}{75.2}  & 91.7  & \multicolumn{1}{c|}{61.0}  & 86.9  &  40.13 & 13.89  & 28.20                       \\ \hline
		\nth{2}                                                                       & \multicolumn{1}{c|}{74.5}  & 91.4  & \multicolumn{1}{c|}{60.4}  & 86.7  &  40.13 & 11.23  & 28.19                       \\ \hline
		\nth{3}                                                                                & \multicolumn{1}{c|}{73.7}  & 90.93 & \multicolumn{1}{c|}{56.9}  & 84.0  &  40.12 & 6.77  &  28.15                      \\ \hline
		\nth{4}                                                                                & \multicolumn{1}{c|}{72.2}  & 89.6  & \multicolumn{1}{c|}{39.8}  & 72.4  &  40.12 & 5.02  &  28.13                      \\ \hline
	\end{tabular}}
	\end{adjustbox}	
\end{table}



\subsection{Effect for Temporal Branch}

\subsubsection{Connect to previous works}
StNet \cite{he2019stnet} design the Temporal block by a single 3D Convolution (Conv-3D) of kernel $3\times1\times1$ to connect the neighbor frames. The temporal block is inserted right after stage \nth{3} and \nth{4} of ResNet backbone. It then introduces the Temporal Xception Block, which includes several Conv-3D of different receptive field and Max-pooling, before passing to a fully-connected layer. Compared to StNet, our temporal branch is more simple and efficient. 

SIFAR \cite{fan2021image} also use a simple approach by stacking the input images along the horizon and vertical axes to create the ``super-image", and then apply the standard image classification on it, such as ResNet101 or Swin-Transformer. In this perspective, SIFAR is the closest work with us. However, our design is more efficient. First, we stack the features after the \nth{2} stage instead of input images. This design can save computation in practice since we can reuse the feature extracted from the early-net during temporal sliding in a long video. Second, we add the dilated convolution to aggregate the temporal information across the frames. More important, this temporal information is computed in global scope and shared between the frames. Finally, SIFAR relies on Swin Transformer backbone, while ViC is simply utilizes all 2D Convolution layers.  
\begin{table}[!]
	\caption{Compare to other close works on Kinetics-400 and Something-Something V2 dataset}
	\label{tab:CloseWork}
	\centering
	\begin{adjustbox}{max width=\columnwidth}
		\small{
			\addtolength{\tabcolsep}{2pt} 
			\begin{tabular}{|l|cc|cc|}
				\hline
				\multirow{2}{*}{Method} & \multicolumn{2}{c|}{Kinetics-400}  & \multicolumn{2}{c|}{SS-v2} \\ \cline{2-5}
				& \multicolumn{1}{c|}{Top-1} & Top-5 & \multicolumn{1}{c|}{Top-1}      & Top-5     \\ \hline
				StNet - R50  \cite{he2019stnet}           & \multicolumn{1}{c|}{69.85} & -     & \multicolumn{1}{c|}{-}          &   -         \\ 
				StNet - R101 \cite{he2019stnet}           & \multicolumn{1}{c|}{71.4}  & -     & \multicolumn{1}{c|}{-}          &   -         \\ \hline
				SIFAR - SwinB - 7  \cite{fan2021image}     & \multicolumn{1}{c|}{79.6}  & 94.4  & \multicolumn{1}{c|}{56.7}       & 83.3      \\
				SIFAR - SwinB - 12 \cite{fan2021image}     & \multicolumn{1}{c|}{\textbf{80}}    & 94.5  & \multicolumn{1}{c|}{60.1}       &  86.8                          \\ \hline
				ViC-S - 3$\times$3                  & \multicolumn{1}{c|}{78.2}  & 93.5  & \multicolumn{1}{c|}{\textbf{64.4}}       & 88.8      \\
				ViC-S - 4$\times$4                  & \multicolumn{1}{c|}{79.5}  & 94.0  & \multicolumn{1}{c|}{\textbf{65.1}}       & 89.6      \\ \hline
		\end{tabular}}
	\end{adjustbox}	
\end{table}

\section{Conclusions}
In this work, we adopt the modernized 2D Convolution network for the task of video action recognition. Our model can compete favorably with other state-of-the-art Vision Transformer while retaining the simplicity and  efficiency of standard 2D ConvNets. Our objective is not to introduce a new or complex temporal module. Instead, we challenge the belief that 2D Convolution is inferior to 3D ConvNet or ViT for action recognition. Therefore, we select a minimal design of the 2D ConvNet, and prove that its performance is still very competitive while preserves the efficiency. We hope that our model can serve as a simple and strong baseline, and offer a practical solution for product deployment.  

{\small
\bibliographystyle{ieee_fullname}
\bibliography{egbib}
}
\appendix
\section{Does the temporal branch actually learn temporal information?}
To answer this question, we evaluate the accuracy when frame's order is shuffled. Should the model learn temporal feature, its performance must be dependent on the frame's temporal structure and the accuracy will be drop in this test. \Tab{Shuffle} shows the behavior of VidConv and several other models under three testing conditions: normal, reverse order and randomly shuffling. We see that, on Kinetics dataset, the accuracy just slightly drops for all models when reverse the order. Uniformer and Swin models get about 3\% accuracy drops when the order is shuffled randomly, while there is almost no dropping for ViC and TimeSformer. On SS-v2 dataset, the effect is more obvious. There are two tendencies. VidConv, TSM and TimeSformer get the worst performance when the order of frames is reverse, while Uniformer and Swin drops most severely if the frame order is random shuffled.   

\begin{table*}
	\centering
	\caption{Evaluate when the frame order is shuffled.}
	\label{tab:Shuffle}
	\begin{tabular}{|c|ccccc||ccccc|}
		\hline
		\multirow{2}{*}{\begin{tabular}[c]{@{}c@{}}Frame \\ Order\end{tabular}} & \multicolumn{5}{c||}{Kinetics K400}                                                                                                                                                                                                                                                                                            & \multicolumn{5}{c|}{Something-Something V2}                                                                                                                                                                                                                                                                                  \\ \cline{2-11} 
		& \multicolumn{1}{c|}{ViC-S} & \multicolumn{1}{c|}{\begin{tabular}[c]{@{}c@{}}TSM \\ R50\end{tabular}} & \multicolumn{1}{c|}{\begin{tabular}[c]{@{}c@{}}TimeS-\\ former\end{tabular}} & \multicolumn{1}{c|}{\begin{tabular}[c]{@{}c@{}}Uni-\\ Former\end{tabular}} & \multicolumn{1}{c||}{Swin-B} & \multicolumn{1}{c|}{ViC-S} & \multicolumn{1}{c|}{\begin{tabular}[c]{@{}c@{}}TSM \\ R50\end{tabular}} & \multicolumn{1}{c|}{\begin{tabular}[c]{@{}c@{}}TimeS-\\ former\end{tabular}} & \multicolumn{1}{c|}{\begin{tabular}[c]{@{}c@{}}Uni-\\ Former-S\end{tabular}} & \multicolumn{1}{c|}{Swin-B} \\ \hline
		normal                                                                  & \multicolumn{1}{c|}{78.2}  & \multicolumn{1}{c|}{71.8}                                                          & \multicolumn{1}{c|}{78}                                                      & \multicolumn{1}{c|}{80.8}                                                  & 80.6                                              & \multicolumn{1}{c|}{64.4}  & \multicolumn{1}{c|}{60.6}                                                       & \multicolumn{1}{c|}{59.5}                                                    & \multicolumn{1}{c|}{67.7}                                                    & 69.6                                              \\ \hline
		reverse                                                                 & \multicolumn{1}{c|}{78.1}  & \multicolumn{1}{c|}{71.7}                                                            & \multicolumn{1}{c|}{78}                                                      & \multicolumn{1}{c|}{80.4}                                                  & 80.2                                              & \multicolumn{1}{c|}{25.4}  & \multicolumn{1}{c|}{19.2}                                                         & \multicolumn{1}{c|}{28.9}                                                    & \multicolumn{1}{c|}{23.7}                                                    & 30.7                                              \\ \hline
		random                                                                  & \multicolumn{1}{c|}{78.1}  & \multicolumn{1}{c|}{70.7}                                                            & \multicolumn{1}{c|}{78}                                                      & \multicolumn{1}{c|}{76.9}                                                  & 77.6                                              & \multicolumn{1}{c|}{40.2}  & \multicolumn{1}{c|}{25.0}                                                         & \multicolumn{1}{c|}{46.0}                                                    & \multicolumn{1}{c|}{16.0}                                                    & 25.0                                              \\ \hline
	\end{tabular}
\end{table*}

We visualize the attention map of VidConv model using CAM \cite{zhou2015cnnlocalization} in Appendix.

\begin{figure*}
	\begin{tabular}{cc}
		\subfloat[break dance]{\includegraphics[width = 0.49\textwidth, height=1.6in]{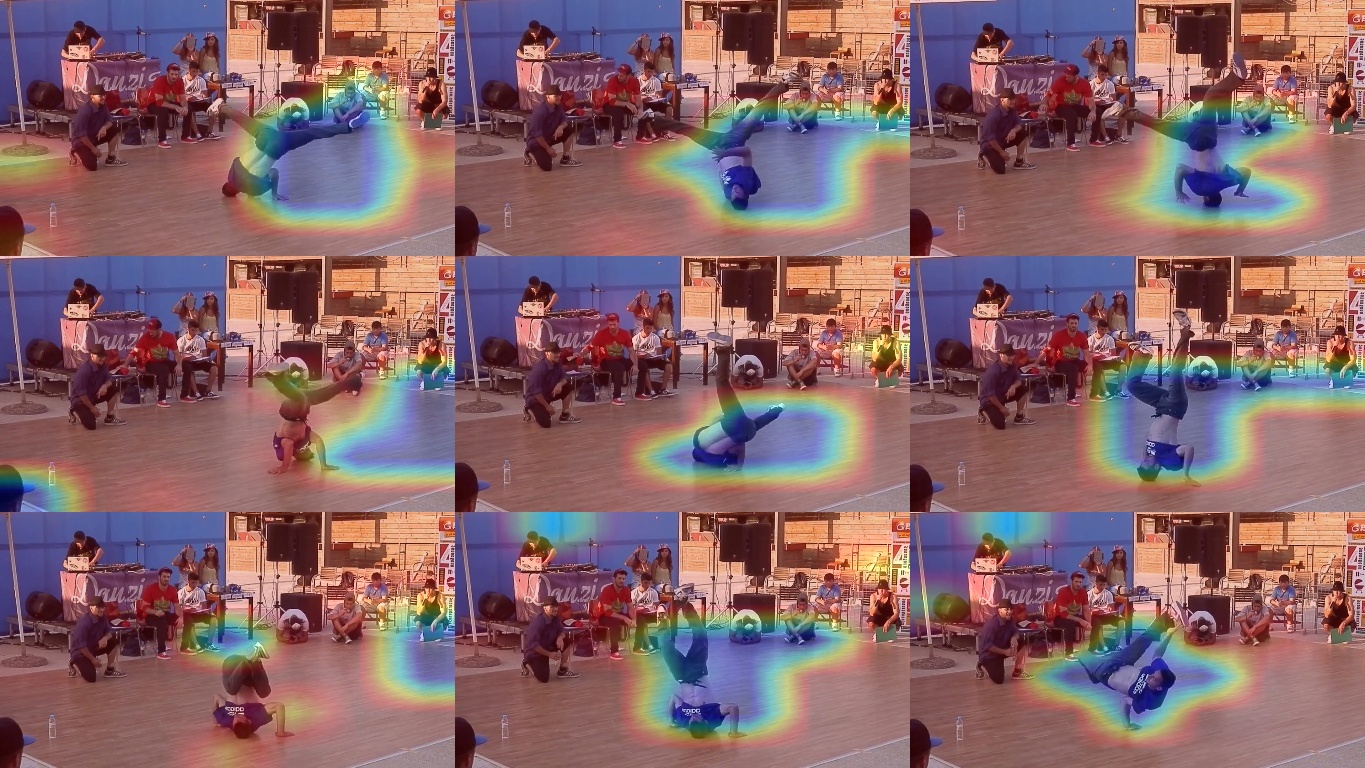}} &
		\subfloat[parasailing]{\includegraphics[width = 0.49\textwidth, height=1.6in]{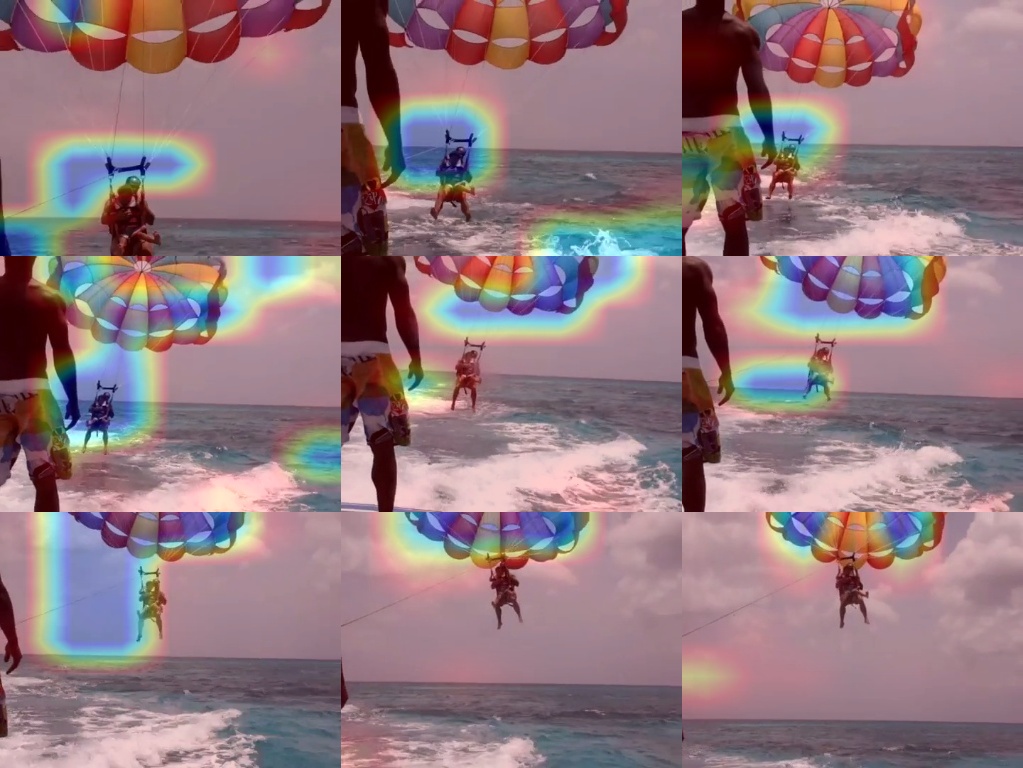}} \\
		\subfloat[long jump]{\includegraphics[width = 0.49\textwidth, height=1.6in]{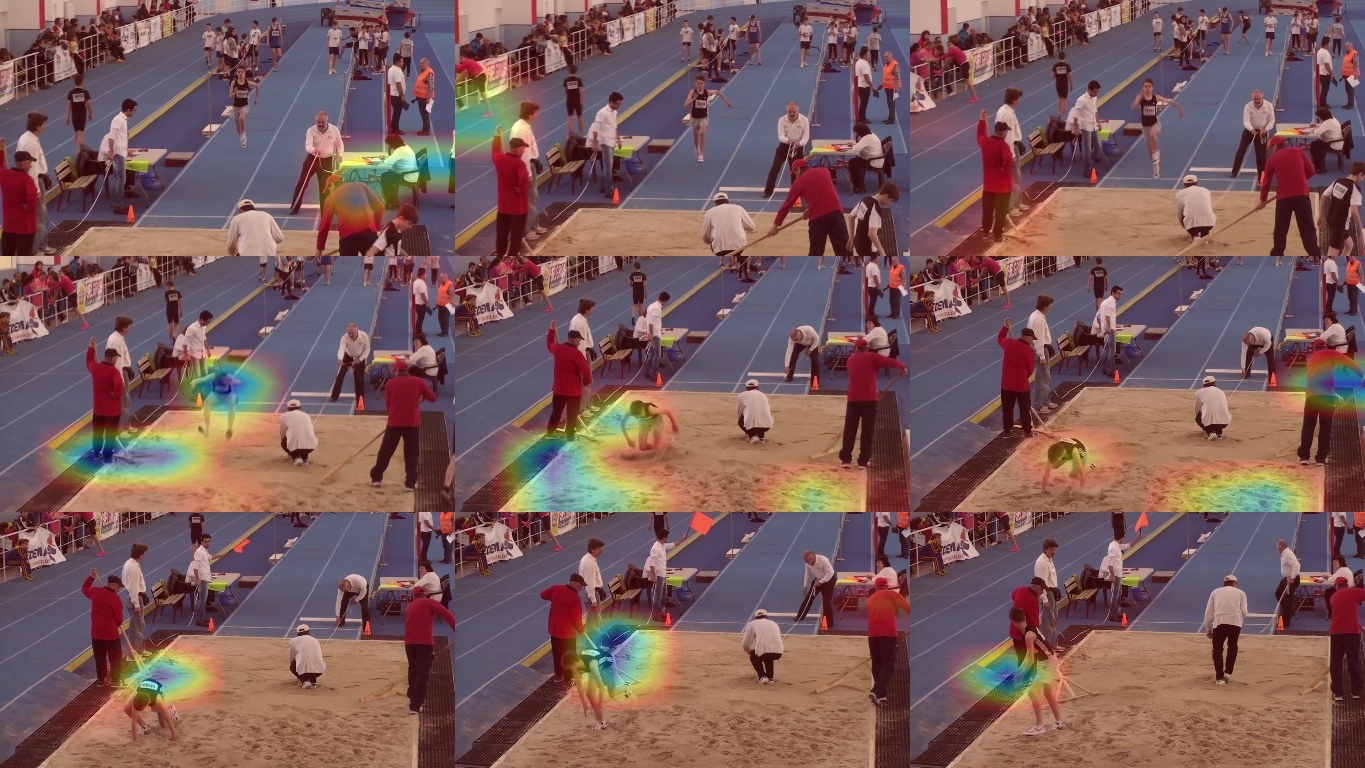}} &
		\subfloat[training dog]{\includegraphics[width = 0.49\textwidth, height=1.6in]{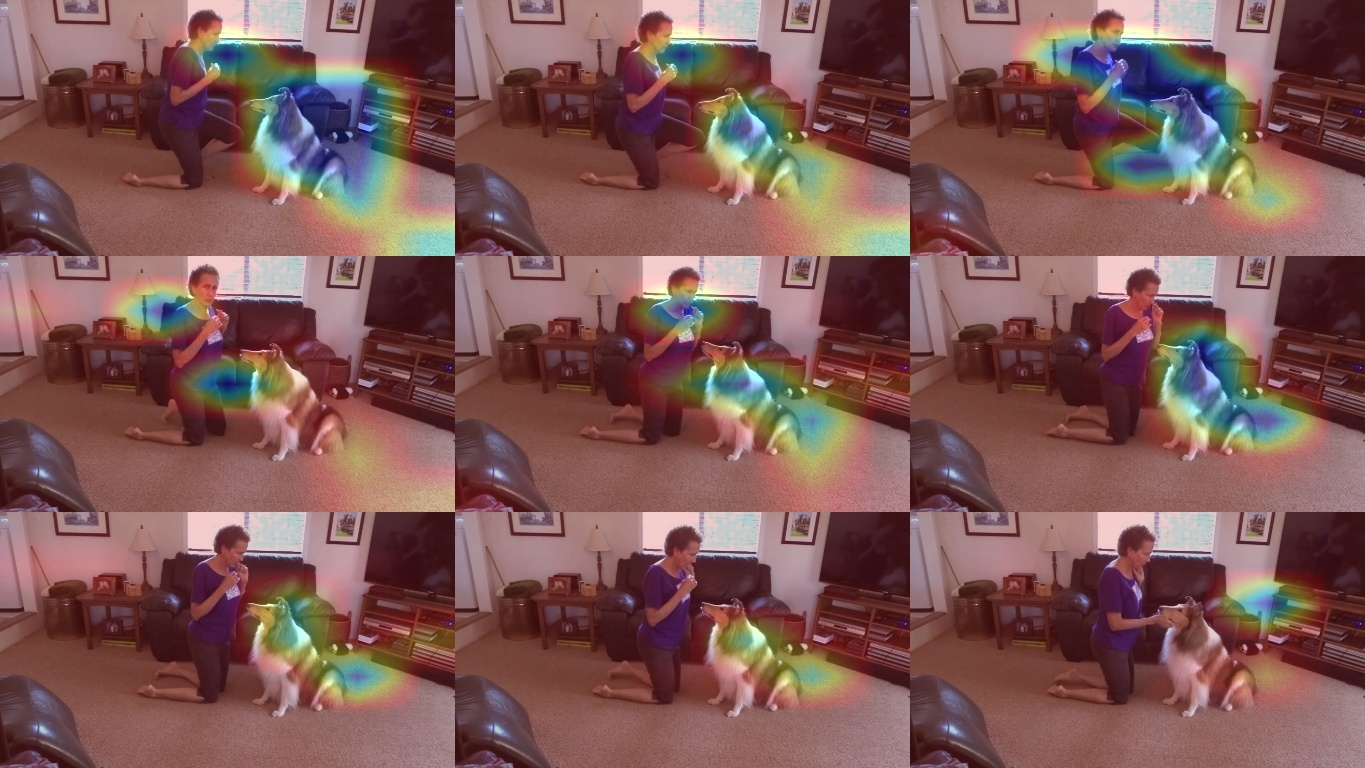}} \\
		\subfloat[making jewelry]{\includegraphics[width = 0.49\textwidth, height=1.6in]{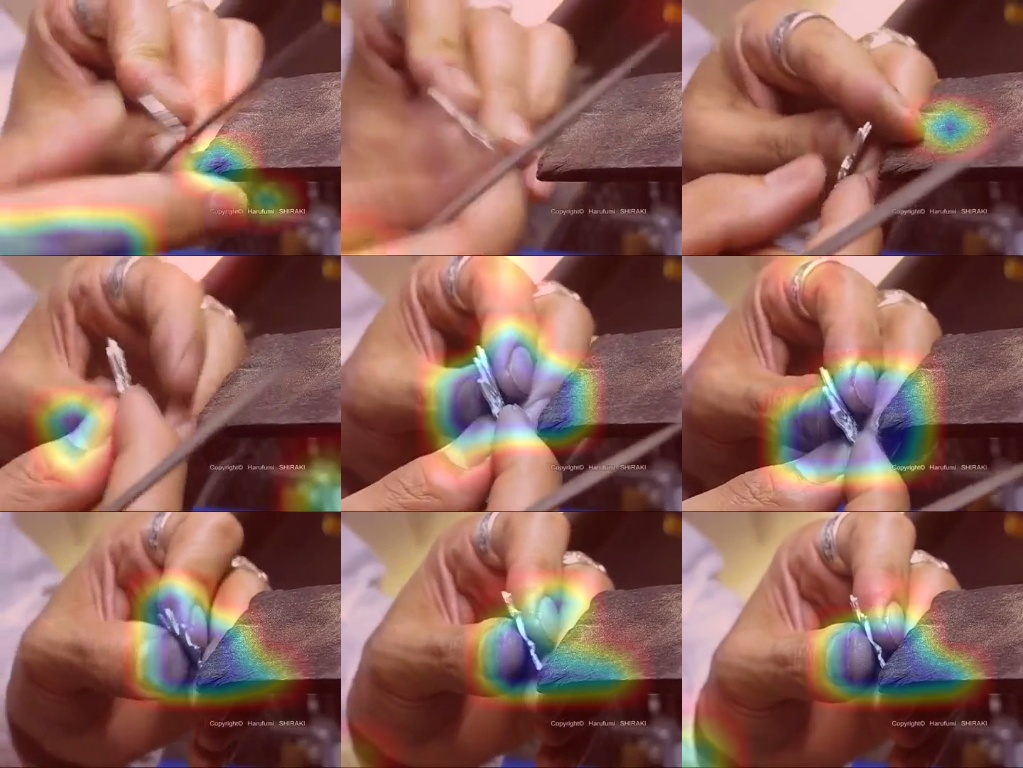}} &
		\subfloat[cracking neck]{\includegraphics[width = 0.49\textwidth, height=1.6in]{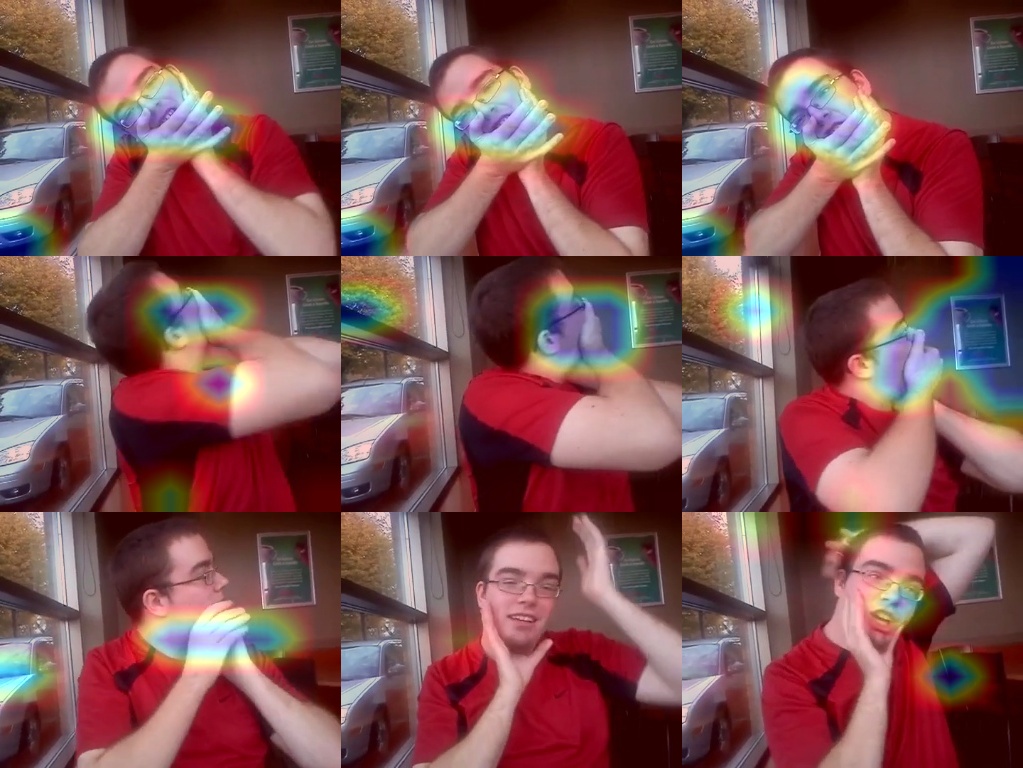}} \\
		\subfloat[sit up]{\includegraphics[width = 0.49\textwidth, height=1.6in]{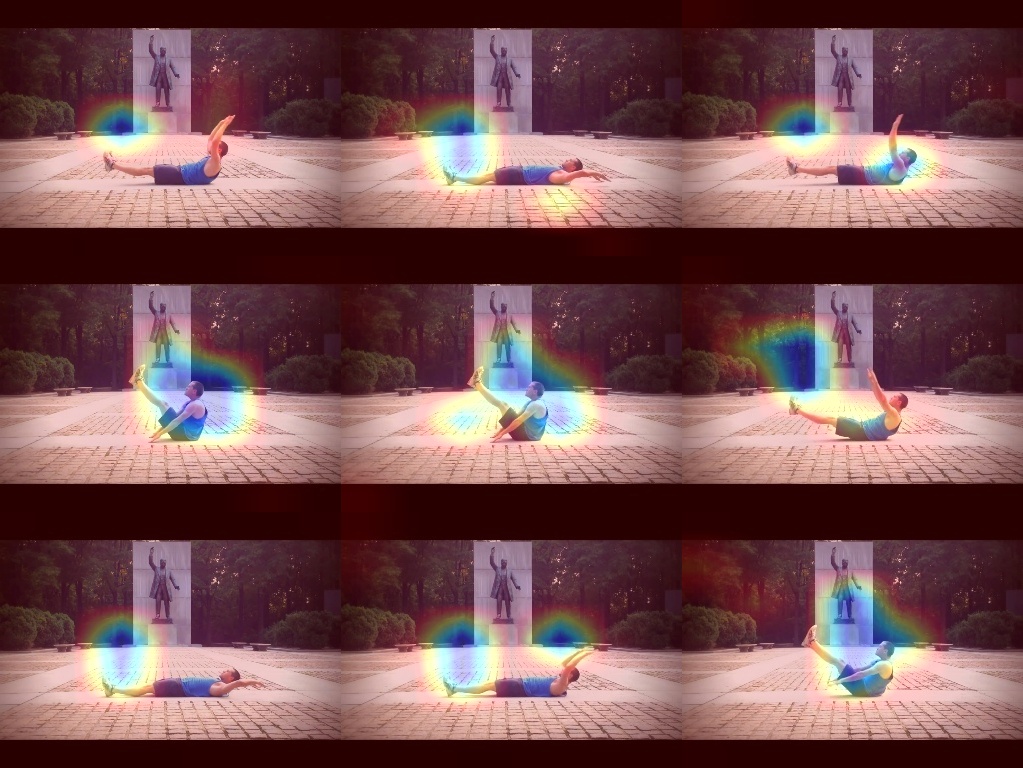}} &
		\subfloat[passing American football (in game)]{\includegraphics[width = 0.49\textwidth, height=1.6in]{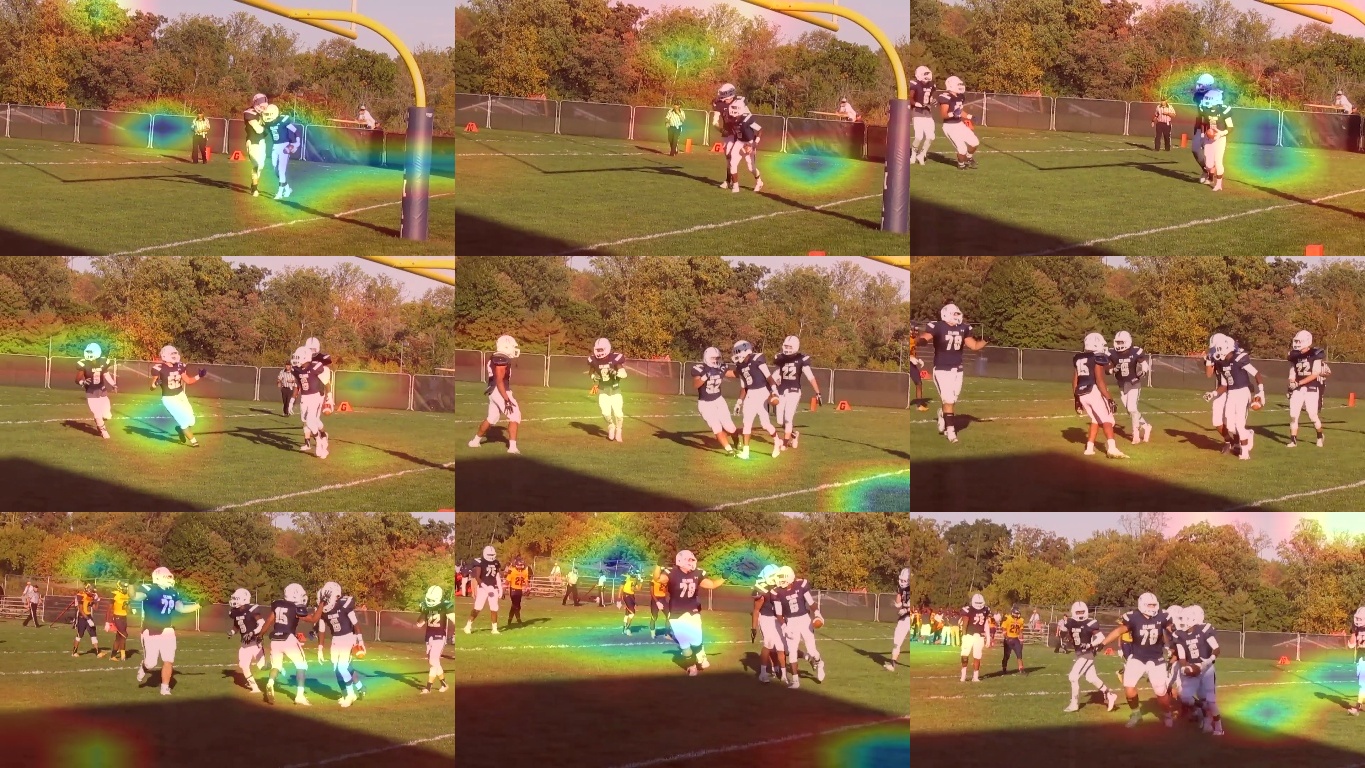}} \\
		\subfloat[slapping]{\includegraphics[width = 0.49\textwidth, height=1.6in]{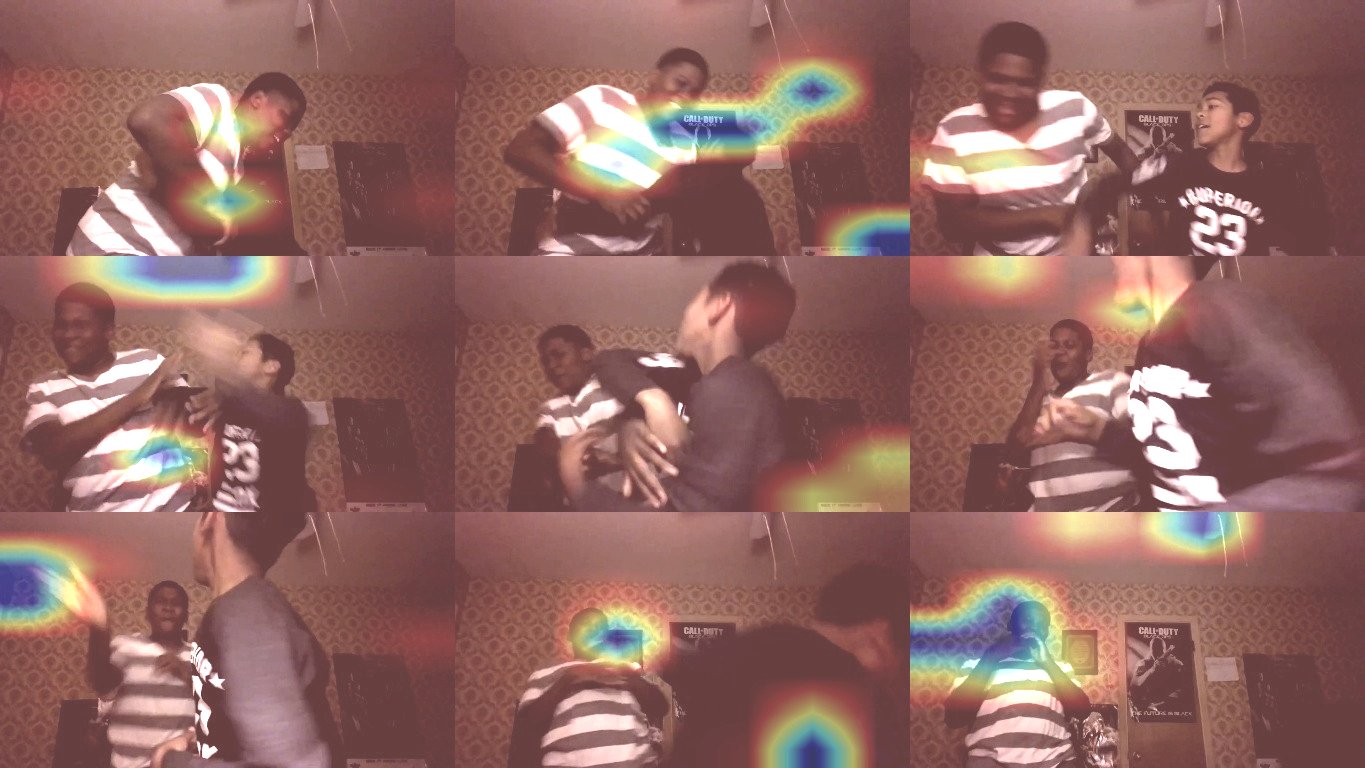}}&
		\subfloat[ice fishing]{\includegraphics[width = 0.49\textwidth, height=1.6in]{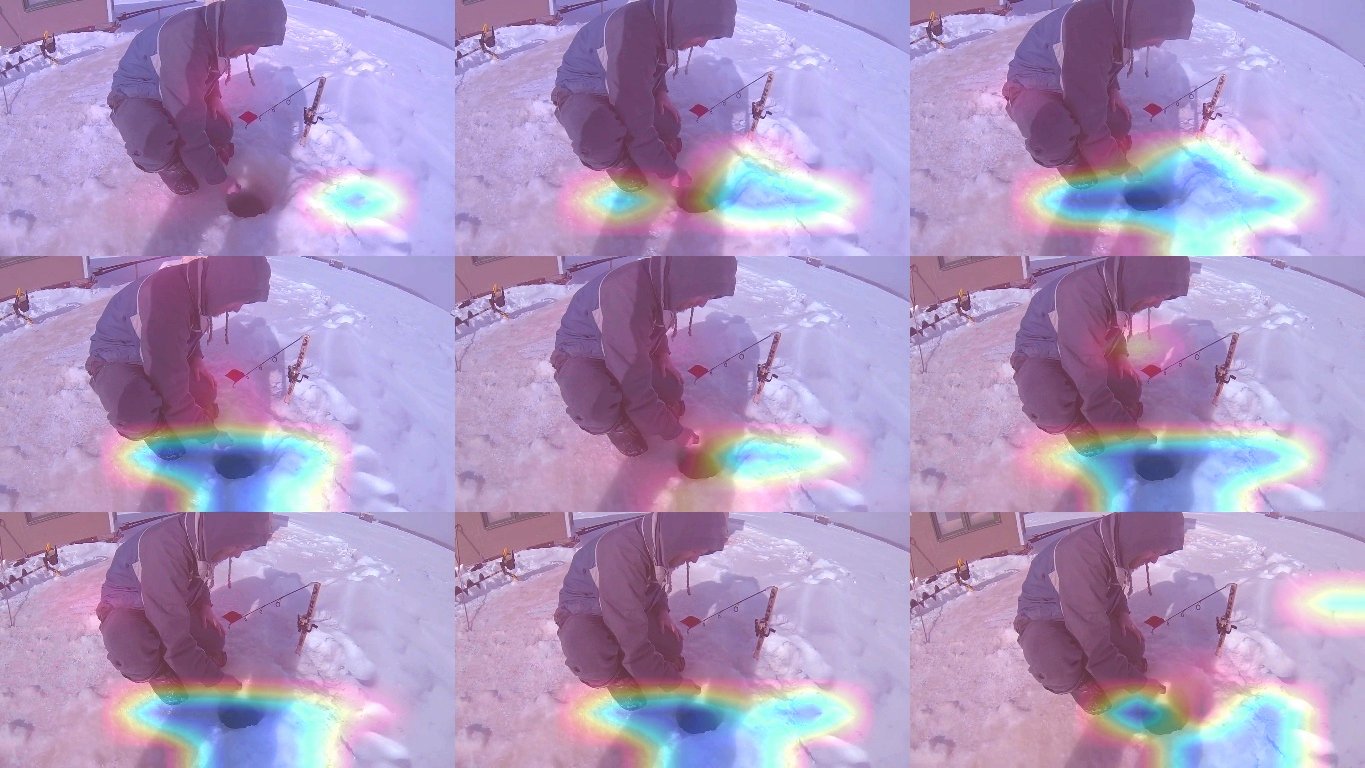}}
	\end{tabular}
	\caption{Visualize the attention areas by Class Activation Map (CAM).}
	\label{fig:CAM}
\end{figure*}

\end{document}